%% file: main_preprint.tex
\documentclass{article}

   \PassOptionsToPackage{numbers, compress}{natbib}
\usepackage[preprint]{neurips_2024}

\usepackage[utf8]{inputenc} % allow utf-8 input
\usepackage[T1]{fontenc}    % use 8-bit T1 fonts
\usepackage{hyperref}       % hyperlinks
\usepackage{url}            % simple URL typesetting
\usepackage{booktabs}       % professional-quality tables
\usepackage{amsfonts}       % blackboard math symbols
\usepackage{nicefrac}       % compact symbols for 1/2, etc.
\usepackage{microtype}      % microtypography
\usepackage{xcolor}         % colors
\usepackage{algorithmic}
\usepackage{graphicx}
\usepackage{textcomp}
\usepackage{quoting}
\usepackage{threeparttable}
\usepackage[subrefformat=parens]{subcaption}
\usepackage{adjustbox}
\usepackage{tabularx,booktabs,ragged2e}
\usepackage{multirow}
\usepackage[normalem]{ulem}
\useunder{\uline}{\ul}{}

\title{Evaluating Robustness of LLMs on Crisis-Related Microblogs across Events, Information Types, and Linguistic Features}

\author{%
Muhammad Imran \\
  Qatar Computing Research Institute\\
  Hamad Bin Khalifa University\\
  Doha, Qatar \\
  \texttt{mimran@hbku.edu.qa}
  \And
  Abdul Wahab Ziaullah \\
  Qatar Computing Research Institute\\
  Hamad Bin Khalifa University\\
  Doha, Qatar \\
  \texttt{awahab@hbku.edu.qa} 
  \And
  Kai Chen\\
  OpenAI\\
San Francisco, CA, USA\\
\texttt{kaichen@openai.com}
\And
    Ferda Ofli \\
  Qatar Computing Research Institute\\
  Hamad Bin Khalifa University\\
  Doha, Qatar \\
  \texttt{fofli@hbku.edu.qa}
}

\begin{document}

\maketitle
\begin{abstract}
  The widespread use of microblogging platforms like X (formerly Twitter) during disasters provides real-time information to governments and response authorities. However, the data from these platforms is often noisy, requiring automated methods to filter relevant information. Traditionally, supervised machine learning models have been used, but they lack generalizability. In contrast, Large Language Models (LLMs) show better capabilities in understanding and processing natural language out of the box. This paper provides a detailed analysis of the performance of six well-known LLMs in processing disaster-related social media data from a large-set of real-world events. Our findings indicate that while LLMs, particularly GPT-4o and GPT-4, offer better generalizability across different disasters and information types, most LLMs face challenges in processing flood-related data, show minimal improvement despite the provision of examples (i.e., shots), and struggle to identify critical information categories like urgent requests and needs. Additionally, we examine how various linguistic features affect model performance and highlight LLMs' vulnerabilities against certain features like typos. Lastly, we provide benchmarking results for all events across both zero- and few-shot settings and observe that proprietary models outperform open-source ones in all tasks.
\end{abstract}

\input{intro}

\input{related_work}
\input{experimental_setup}
\input{results}
\input{conclusion}
\bibliographystyle{abbrv}
\bibliography{bibliography}

\appendix

\section{Class-wise results}
\label{appendix:class-results}
Tables~\ref{tab:gpt_models} and \ref{tab:llama_models} present the class-wise results across various shots for all the models. %proprietary models (GPT-3.5, GPT-4, and GPT-4o) and open-source models (Llama-2 13B, Llama-3 8B, and Mistral 7B), respectively.

\begin{table}[h]
\centering
\caption{F1 Scores for proprietary models (GPT-3.5, GPT-4, and GPT-4o) across classes and k-shots}
\label{tab:gpt_models}
\resizebox{13cm}{!}{%
\begin{tabular}{l|ccccc|ccccc|ccccc}
\toprule
\textbf{Class} & \multicolumn{5}{c|}{\textbf{GPT-3.5}} & \multicolumn{5}{c|}{\textbf{GPT-4}} & \multicolumn{5}{c}{\textbf{GPT-4o}} \\
\cmidrule{2-16}
 & ZS & 1S & 3S & 5S & 10S & ZS & 1S & 3S & 5S & 10S & ZS & 1S & 3S & 5S & 10S \\
\midrule
Caution and advice & 0.65 & 0.84 & 0.89 & 0.86 & 0.87 & 0.91 & 0.89 & 0.90 & 0.88 & 0.89 & 0.81 & 0.90 & 0.88 & 0.86 & 0.84 \\
Rescue volunteering & 0.89 & 0.95 & 0.94 & 0.94 & 0.95 & 0.92 & 0.92 & 0.90 & 0.88 & 0.88 & 0.91 & 0.93 & 0.93 & 0.91 & 0.89 \\
Requests or urgent needs & 0.68 & 0.58 & 0.58 & 0.50 & 0.55 & 0.71 & 0.66 & 0.74 & 0.68 & 0.73 & 0.70 & 0.75 & 0.75 & 0.81 & 0.85 \\
Infrastructure damage & 0.91 & 0.77 & 0.79 & 0.78 & 0.80 & 0.85 & 0.79 & 0.80 & 0.70 & 0.75 & 0.84 & 0.84 & 0.79 & 0.81 & 0.81 \\
Sympathy and support & 0.85 & 0.83 & 0.79 & 0.79 & 0.79 & 0.91 & 0.89 & 0.88 & 0.90 & 0.88 & 0.91 & 0.91 & 0.91 & 0.90 & 0.90 \\
Injured or dead people & 0.80 & 0.88 & 0.86 & 0.84 & 0.83 & 0.91 & 0.88 & 0.91 & 0.85 & 0.85 & 0.94 & 0.94 & 0.93 & 0.93 & 0.91 \\
Displaced people & 0.83 & 0.74 & 0.68 & 0.62 & 0.66 & 0.90 & 0.74 & 0.78 & 0.62 & 0.81 & 0.88 & 0.73 & 0.74 & 0.74 & 0.80 \\
Missing or found people & 0.87 & 0.81 & 0.86 & 0.82 & 0.85 & 0.85 & 0.88 & 0.84 & 0.83 & 0.81 & 0.82 & 0.88 & 0.89 & 0.90 & 0.93 \\
Not humanitarian & 0.46 & 0.78 & 0.72 & 0.77 & 0.77 & 0.70 & 0.88 & 0.80 & 0.91 & 0.81 & 0.91 & 0.86 & 0.88 & 0.88 & 0.87 \\
\bottomrule
\end{tabular}%
}
\end{table}

\begin{table}[h]
\centering
\caption{F1 Scores for open-source models (Llama-2 13B, Llama-3 8B, and Mistral 7B) across classes and k-shots}
\label{tab:llama_models}
\resizebox{13cm}{!}{%
\begin{tabular}{l|ccccc|ccccc|ccccc}
\toprule
\textbf{Class} & \multicolumn{5}{c|}{\textbf{Llama-2 13B}} & \multicolumn{5}{c|}{\textbf{Llama-3 8B}} & \multicolumn{5}{c}{\textbf{Mistral 7B}} \\
\cmidrule{2-16}
 & ZS & 1S & 3S & 5S & 10S & ZS & 1S & 3S & 5S & 10S & ZS & 1S & 3S & 5S & 10S \\
\midrule
Caution and advice & 0.61 & 0.91 & 0.82 & 0.47 & - & 0.52 & 0.62 & 0.62 & 0.58 & 0.63 & 0.36 & 0.50 & 0.71 & 0.88 & 0.36 \\
Rescue volunteering & 0.75 & 0.53 & 0.35 & 0.42 & - & 0.50 & 0.47 & 0.50 & 0.44 & 0.35 & 0.95 & 0.96 & 0.87 & 0.65 & 0.82 \\
Requests or urgent needs & 0.73 & 0.78 & 0.84 & 0.83 & - & 0.34 & 0.28 & 0.27 & 0.25 & 0.21 & 0.56 & 0.55 & 0.66 & 0.70 & 0.63 \\
Infrastructure damage & 0.77 & 0.59 & 0.44 & 0.64 & - & 0.54 & 0.46 & 0.34 & 0.24 & 0.30 & 0.83 & 0.73 & 0.45 & 0.54 & 0.53 \\
Sympathy and support & 0.90 & 0.93 & 0.92 & 0.90 & - & 0.75 & 0.70 & 0.74 & 0.73 & 0.64 & 0.83 & 0.83 & 0.79 & 0.86 & 0.63 \\
Injured or dead people & 0.66 & 0.75 & 0.65 & 0.82 & - & 0.79 & 0.78 & 0.77 & 0.75 & 0.74 & 0.88 & 0.79 & 0.60 & 0.49 & 0.52 \\
Displaced people & 0.58 & 0.20 & 0.24 & 0.47 & - & 0.36 & 0.34 & 0.22 & 0.27 & 0.30 & 0.69 & 0.36 & 0.46 & 0.66 & 0.68 \\
Missing or found people & 0.70 & 0.78 & 0.80 & 0.79 & - & 0.47 & 0.34 & 0.48 & 0.51 & 0.44 & 0.64 & 0.69 & 0.65 & 0.75 & 0.77 \\
Not humanitarian & 0.34 & 0.73 & 0.69 & 0.56 & - & 0.53 & 0.57 & 0.53 & 0.57 & 0.50 & 0.69 & 0.86 & 0.87 & 0.60 & 0.89 \\
\bottomrule
\end{tabular}%
}
\end{table}

\end{document}

%% file: intro.tex
\section{Introduction}
Microblogging platforms like X (formerly Twitter) are vital during large-scale disasters~\cite{vieweg2010microblogging}. They facilitate real-time communication for the public to share firsthand experiences, report damage to infrastructure, and most importantly, seek assistance~\cite{mihunov2020use,alam2018twitter}. Moreover, local governments are increasingly leveraging these non-traditional data sources to enhance their situational awareness and quickly identify humanitarian needs, %such as medical aid, shelter, and food
 and inform their response strategies accordingly~\cite{splendiani2022crisis,landwehr2016using}.

Despite their accessibility, data from social media platforms are often highly noisy~\cite{he2017signals}. During large-scale disasters, the volume of messages can reach millions per day, filled with irrelevant content and chatter~\cite{castillo2016big}. This deluge makes it challenging for local authorities to identify reports critical for humanitarian response. %This deluge makes it challenging for local response authorities to identify urgent, time-critical reports. 
Previous research has addressed this issue by developing supervised machine learning models that filter through the raw data to identify relevant information~\cite{imran2015processing,olteanu2014crisislex}. However, these models typically struggle with generalizability across different disasters or geographic locations due to the problems of domain shift~\cite{geonet, rebu_2}. Techniques like domain adaptation or transfer learning have been proposed to alleviate these challenges~\cite{gama2014survey, TL_in_RS_RSE}. Nonetheless, when the categories of interest change, training new machine learning models becomes necessary. This process requires human-labeled data, which is time-intensive, and can slow down response efforts.

Large Language Models (LLMs) demonstrate a strong capability to comprehend natural language and generalize across various NLP tasks~\cite{yang2024harnessing}. Despite numerous studies assessing LLMs' effectiveness with well-structured web data~\cite{zhao2023survey} and noisy social media content, mainly in non-humanitarian context~\cite{zhang2024toward,ignat2024cross}, no previous research presents a thorough analysis of their robustness in processing disaster-related social media data. In this paper, we present a comprehensive analysis of social media data collected from 19 major disasters across multiple countries using six well-known LLMs, including GPT-3.5~\cite{brown2020language}, GPT-4~\cite{achiam2023gpt}, GPT-4o~\cite{gpt4o}, Llama-2 13B~\cite{touvron2023llama}, Llama-3 8B~\cite{dubey2024llama}, and Mistral 7B~\cite{jiang2023mistral}. We assess the effectiveness of these proprietary and open-source LLMs in handling different disaster types and information categories, and their performance with data from both native and non-native English-speaking countries. We also examine how various linguistic features influence LLMs' performance. Additionally, our study provides benchmarking results for each of the 19 disaster events and evaluates the overall model performance in both zero- and few-shot settings.

Our findings indicate that proprietary models (i.e., GPT-4 and GPT-4o) generally outperform open-source models (i.e., Llama-2 13B, Llama-3 8B, and Mistral 7B) on various tasks. %,---which is consistent with previous studies. 
However, GPT models notably struggle with processing data from flood incidents. Moreover, certain information types, such as \textit{requests or urgent needs}, consistently challenge all models, with all GPTs achieving F1 below 0.60. Open-source models also display weaknesses in handling classes like \textit{caution and advice} and \textit{requests or urgent needs}. Additionally, we find that providing models with class-specific examples does not generally enhance their performance. 

The rest of the paper is organized as follows. We summarize the related work in Section~\ref{sec:related_work}, describe our assessment methodology in Section~\ref{sec:asmfrwk}, and present our results and discussions in Section~\ref{sec:results_discussion}. Finally, we conclude the paper and provide a future work plan in Section~\ref{sec:conclusion}.

%% file: related_work.tex
\section{Related Work}
\label{sec:related_work}

In crisis informatics literature, several studies introduced large-scale crisis-related microblog datasets and presented baseline results using both classical machine learning algorithms (e.g., Random Forest, Support Vector Machines, etc.) as well as deep learning models (e.g., RNNs, LSTMs, CNNs, etc.)~\cite{olteanu2014crisislex, imran2016lrec}. Later, researchers undertook an effort to consolidate available datasets and tasks for benchmarking transformer-based models such as BERT~\cite{devlin2019bert}, DistilBERT~\cite{sanh2020distilbert} and RoBERTa~\cite{Liu2019RoBERTaAR}, and showed that the transformer-based models typically outperform~\cite{alam2021crisisbench}. %ed CNNs and other DL algorithms on various performance indicators~\cite{alam2021crisisbench}. 
A more comprehensive crisis-related dataset along with benchmarking results were presented in \cite{alam2021humaid}. %for various disaster types such as earthquakes, hurricanes, floods and wildfires. 
Likewise, a more recent study~\cite{wilkho2024ff} presented a BERT-based ensemble model, FF-BERT, for the classification of flash flooding messages. Their evaluations examined various BERT-based ensemble models on a specially curated dataset of 21,180 paragraphs of text. Meanwhile, \cite{han2024quakebert} developed QuakeBERT and showed better performance to assess physical and social impacts of an earthquake through microblogs. %Their model outperformed other classical ML models.% on relevance and classification of earthquake-related microblogs.

Previous research has shown that transformer-based models outperform traditional ML algorithms on various metrics. Recent efforts have focused on using more powerful LLMs across diverse fields and tasks. For instance, LLMeBench has assessed LLMs on multiple NLP tasks such as sentiment analysis and summarization \cite{dalvi2023llmebench}. Additionally, studies like \cite{ziaullah2024monitoring} have applied LLMs to crisis-related tasks, evaluating models like Mistral 7B \cite{jiang2023mistral} for their ability to analyze disaster-related tweets. Further, Llama-2 and Mistral have been fine-tuned for disaster response guidance, as presented in \cite{otal2024llm}. This paper builds upon these findings by analyzing LLMs on a crisis-related dataset, exploring LLMs' performance across various disaster types, information types, and the linguistic features of the messages, to identify their capabilities and weaknesses.% of LLMs in crisis microblog classification.
%In this paper, we extend the evaluation of LLMs classification on crisis-related dataset with thorough analysis on the nature of the disaster, the type of event, the linguistic peculiarities of the messages, etc. Through this work we aim to highlight the capabilities and weaknesses of state-of-the-art LLMs on crisis-related microblog classification.

% . A study on the use of open-source LLM; Mistral 7B, for its capabilities in extracting severity and damage from disaster tweets was conducted in \cite{ziaullah2024monitoring}. Here, a mixture of synthetic and real tweets were provided to decode the damage and severity on critical infrastructure facilities during a natural disaster. 

% All these studies highlight additional need for improved models that are capable for addressing several classification tasks across several different disasters. Through this study we address such need by analysing well known open-source and proprietary Large Language Models on well-known disaster dataset. 

\begin{figure*}[t]
        \centering
            \includegraphics[width=\textwidth]{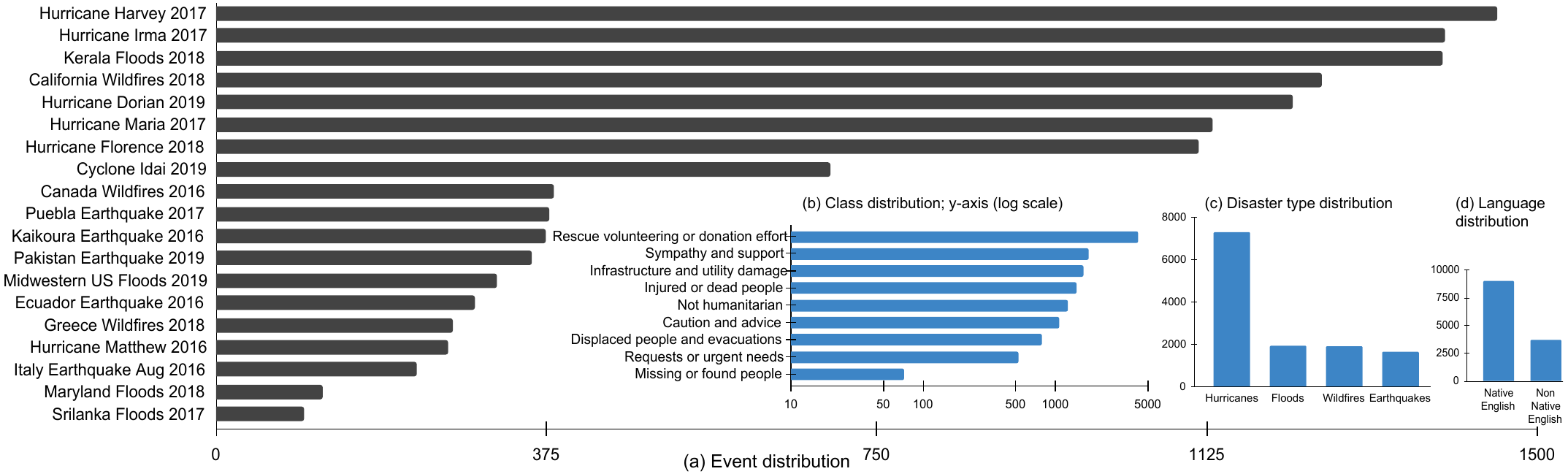}
            % \caption{Data Distribution in Revised HumAID Dataset (a) (Upper-Left) Labels Distribution (b) (Upper-Middle) Language-Type Distribution (c) (Upper-Right) Disaster-Type Distribution
            % (d) (Lower) Event Distribution}%           
         \caption{Data distributions for (a) events, (b) information types/classes, (c) disaster types, and (d) native/non-native English countries} 
        \label{fig:all_dist}
\end{figure*}

%% file: experimental_setup.tex
\section{Assessment Methodology}
\label{sec:asmfrwk}

The increasing complexity and frequency of natural disasters worldwide necessitate AI models, particularly LLMs, that can effectively generalize across various types of disasters (e.g., floods, earthquakes, etc.), languages (English vs.\ Non-English), and different types of information shared on social media (e.g., warnings, urgent needs, damage reports, etc.). We evaluate the performance of LLMs in processing social media content from different types of disasters in countries that use different languages. Additionally, we investigate how open-source and proprietary models differ in performance and assess the role of few-shot learning, where LLMs are provided with examples, on their effectiveness. For this purpose, this paper addresses the following four key questions. 

%\paragraph{Research questions:}
\begin{enumerate}
\item How do LLMs perform for different types of natural disasters (e.g., floods, wildfires)?
%\item What is the capability of LLMs in interpreting various types of social media information at the onset of a disaster?
\item What is LLMs' ability to interpret various types of social media data during disasters? %What is the capability of LLMs in interpreting various types of social media information during disasters?
%\item How does the performance of LLMs vary for countries where the native language is not English?
\item How do LLMs perform for countries where the native language is not English?% in non-English speaking countries?
%\item Are there specific linguistic features or sentence structures that significantly affect the performance of LLMs?
\item Do certain linguistic features or sentence structures significantly impact LLMs performance?
\end{enumerate}

\subsection{Dataset and Models}
\label{sec:Dataset}

% \begin{figure*}[!htbp]
%         \centering
%             \includegraphics[width=0.97\textwidth]{figs/data_distributions_v5.pdf}
%             % \caption{Data Distribution in Revised HumAID Dataset (a) (Upper-Left) Labels Distribution (b) (Upper-Middle) Language-Type Distribution (c) (Upper-Right) Disaster-Type Distribution
%             % (d) (Lower) Event Distribution}%           
%          \caption{Data distributions for (a) events, (b) information types/classes, (c) disaster types, and (d) native/non-native English countries} 
%         \label{fig:all_dist}
%         \Description[]{}
% \end{figure*}

To answer our research questions, we utilize HumAID~\cite{alam2021humaid} dataset comprising 77,196 tweets from 19 different natural disasters that occurred in 11 distinct countries (3 native English-speaking and 8 non-English-speaking) between 2016 to 2019. The tweets in the dataset are labeled by paid crowdsourcing workers into ten distinct information categories (acronyms): \textit{(1) caution and advice (CA)---(2) sympathy and support (SS)---(3) requests or urgent needs (RUN)---(4) displaced people and evacuations (DPE)---(5) injured or dead people (IDP)---(6) missing or found people (MFP)---(7) infrastructure and utility damage (IUD)---(8) rescue volunteering or donation effort (RVDE)---(9) other relevant information (ORI)---} and \textit{(10) not humanitarian (NH)}. We drop the \textit{``other relevant information''} class from our analysis as it mainly contains general event-related information that does not belong to other categories. The dataset is already split into train, development, and test sets. We use the test split (N=15,160) for our experiments. Figure~\ref{fig:all_dist} shows various distributions of our dataset.

\paragraph{Models} We select six well-known LLMs (three proprietary and three open-source) for this study. We choose GPT-3.5~\cite{brown2020language}, GPT-4~\cite{achiam2023gpt}, and GPT-4o~\cite{gpt4o} from OpenAI as our proprietary models and Llama-2 13B~\cite{touvron2023llama}, Llama-3 8B~\cite{dubey2024llama} and Mistral 7B~\cite{jiang2023mistral} as our open-source models. All six models are known for their language understanding capabilities across various NLP tasks.

%This study focuses on both proprietary and open-source LLMs that are well known for their respective performances in various benchmarks \cite{minaee2024large}. For this study we specifically chose such distinction to determine if the benefits of proprietary models outweigh their cost in classification of disaster-related messages compare to open-source models. To this end, we selected \AW{a well-known opensource model from Meta i.e. LLaMA-2 13B \cite{touvron2023llama} and its competitor Mistral 7B \cite{jiang2023mistral} from Mistral AI}, while GPT 3.5 and GPT 4 \AW{ are proprietary models from OpenAI and are current state-of-the-art } as proprietary models. \AW{In general Mistral 7B has shown to outperform LLaMA-2 13B over several benchmarks, while GPT 4 is the next iteration from GPT 3.5, so we anticipate Mistral 7B and GPT 4 to outperform LLaMA-2 13B and GPT 3.5, respectively, in this benchmarking as well. Moreover, both LLaMA-2 and Mistral 7B had been previously used for monitoring critical infrastructure facilities during natural disaster via classification of social media messages \cite{ziaullah2024monitoring}, it is therefore quite pertinent to have them benchmarked on disaster dataset such as HumAID. } 

\begin{figure*}[t]
  \centering
    \subfloat[Proprietary models]
    {\includegraphics[width=.46\linewidth]{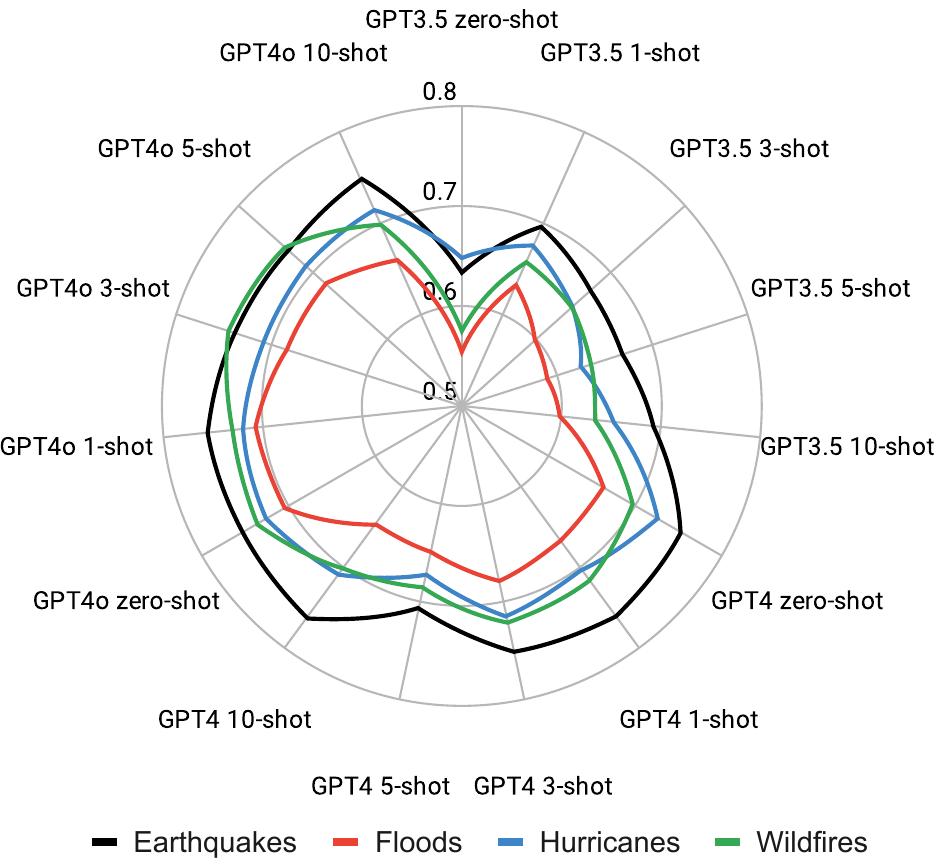}}\hspace{1cm}%\hfill
    %\subfloat[Open-source models] {\includegraphics[width=.44\linewidth]{figs/Spider_Disaster_OS.pdf}}
    \subfloat[Open-source models] {\includegraphics[width=.46\linewidth]{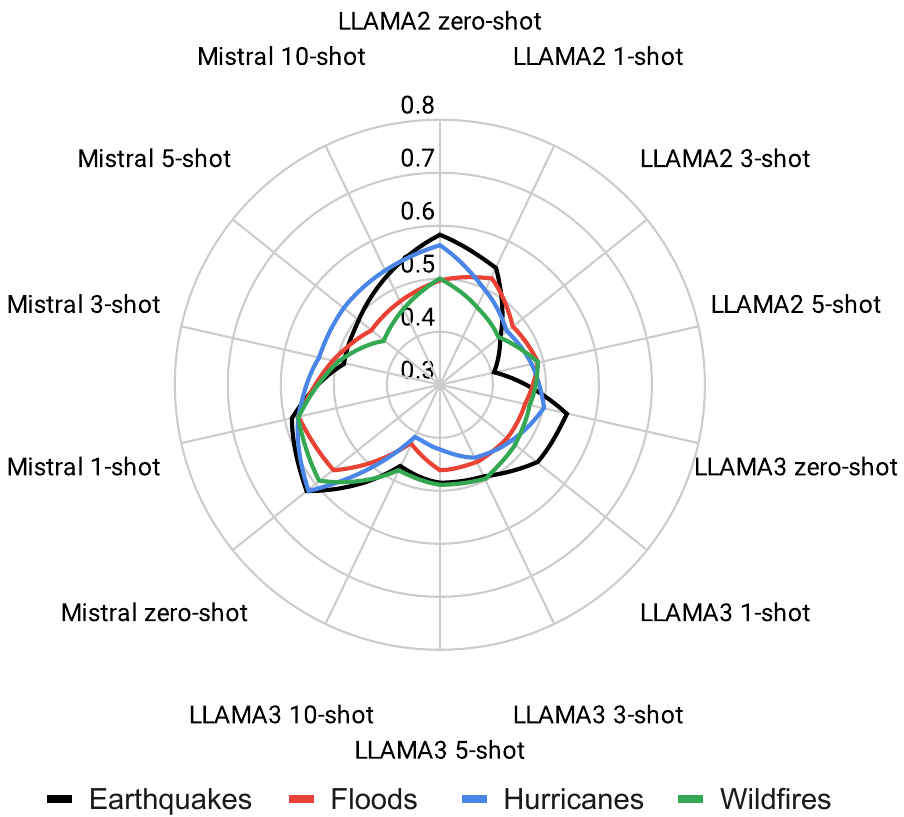}}
  \caption{Performance (F1-scores) of LLMs across disaster types and few-shot settings}
  \label{fig:Disaster_wise}
\end{figure*}

\subsection{Experimental Design}
We evaluate the LLMs for the classification task in two settings: zero-shot and few-shot. In the zero-shot setting, the models operate without any class-specific examples, relying solely on their pre-trained capabilities to perform the task. In the few-shot setting, models receive examples for each class to improve class-specific performance. For instance, in a three-shot experiment, we provide the model with three carefully selected tweets per class from the training set, totaling 30 examples for ten classes. For all experiments, we set the temperature parameter to zero. We use the following prompt across all experiments, except for Llama-2 and Mistral, where we provide additional instructions to control for verbosity. %Full prompts, shots, data, and code are provided in the supplementary material.

\smallskip
\noindent {\footnotesize{
{\bf Prompt:} ``Read the category names and their definitions below, then classify the following tweet into the appropriate category. In your response, mention only the category name.

~Category name: category definition\\
\it{
- Caution and advice: Reports of warnings issued or lifted, guidance and tips related to the disaster.\\
- Sympathy and support: Tweets with prayers, thoughts, and emotional support.\\
- Requests or urgent needs: Reports of urgent needs or supplies such as food, water, clothing, money,...\\% medical supplies or blood.\\
- Displaced people and evacuations: People who have relocated due to the crisis, even for a short time...\\% (includes evacuations).\\
- Injured or dead people: Reports of injured or dead people due to the disaster.\\
- Missing or found people: Reports of missing or found people due to the disaster.\\
- Infrastructure and utility damage: Reports of any type of damage to infrastructure such as buildings, houses,...\\% roads, bridges, power lines, communication poles, or vehicles.\\
- Rescue volunteering or donation effort: Reports of any type of rescue, volunteering, or donation efforts...\\% such as people being transported to safe places, people being evacuated, people receiving medical aid or food, people in shelter facilities, donation of money, or services, etc.\\
%- Other relevant information: If the tweet does not belong to any of the above categories, but it still contains important information useful for humanitarian aid, belong to this category.\\
- Not humanitarian: If the tweet does not convey humanitarian aid-related information."

\noindent Tweet: \{input tweet\}\\
Category:
}}}

%% file: results.tex
\section{Results and Discussion}
\label{sec:results_discussion}
%In this section, we present and discuss our results. All results are reported as F1-scores, which represent the harmonic mean of precision and recall. Due to the input token size limitations, we could not execute Llama-2 10-shot experiment.

%\AW{To address the research question presented in Section \ref{sec:asmfrwk}, we analyse the overall performance of both open-source and proprietary LLMs on HumAID dataset in Section \ref{sec:overall} as well as observe trends of performance variations by utilizing few-shots in the prompts. We then compare the performance of these LLMs on different disaster types and event types in Sections \ref{sec:disaster-wise} \& \ref{sec:event-wise}, respectively. Analysis of LLMs performance on various information-types or classes is presented in Section \ref{sec:class-wise}. The effects of language usage in various geographical regions of the disaster and the effects of various language features is discussed in Section \ref{sec:language-analysis}.   }
%
% In this work we distinguished LLMs based on proprietary and open-source models for various comparisons which include overall performance,  performance against various disaster types and event types, performance against various class labels, benefits of utilizing few-shots, and analysing based on various language patterns. 
%
 %We therefore evaluated Llama-2 on zero, one, three and five shot prompts. 

% Although, despite not being too exhaustive, these unique evaluation dimensions provide valuable insights onto to potential areas where these models can be further improved. 

\subsection{Disaster Type Analysis}
\label{sec:disaster-wise}
Our first research question examines how LLMs perform across different types of disasters. We analyze data from 19 events, grouped into four event types: 5 earthquakes, 7 hurricanes, 3 wildfires, and 4 floods. We present results for both proprietary and open-source models and compare their performance in zero-shot and few-shot (i.e., 1, 3, 5, and 10) settings.
% Our dataset consists of four disaster types (i.e., earthquakes, hurricanes, wildfires, and floods). Each of these disaster types consist of distinct vocabulary synonymous to the nature of impacts they bring as well as methods of humanitarian responses. We therefore, grouped the dataset based on the disaster type to determine if a pattern emerges. We additionally distinguish how proprietary models  perform against open-source models \AW{in handling various disaster types, by separately plotting their F1-scores grouped by disaster types in Figure \ref{fig:Disaster_wise}}. \AW{From the fig. we see a clear pattern of performance trends on different disaster types emerges for proprietary models while open-source models show a mixed behaviour in handling different disaster types.}  

\begin{figure*}[t]
  \centering
    \subfloat[GPT-4o in zero- and few-shot settings]
    %{\includegraphics[width=.35\linewidth]{new_figs/Spider_Class_GPTs.pdf}}\hspace{1cm}
    {\includegraphics[width=.35\linewidth]{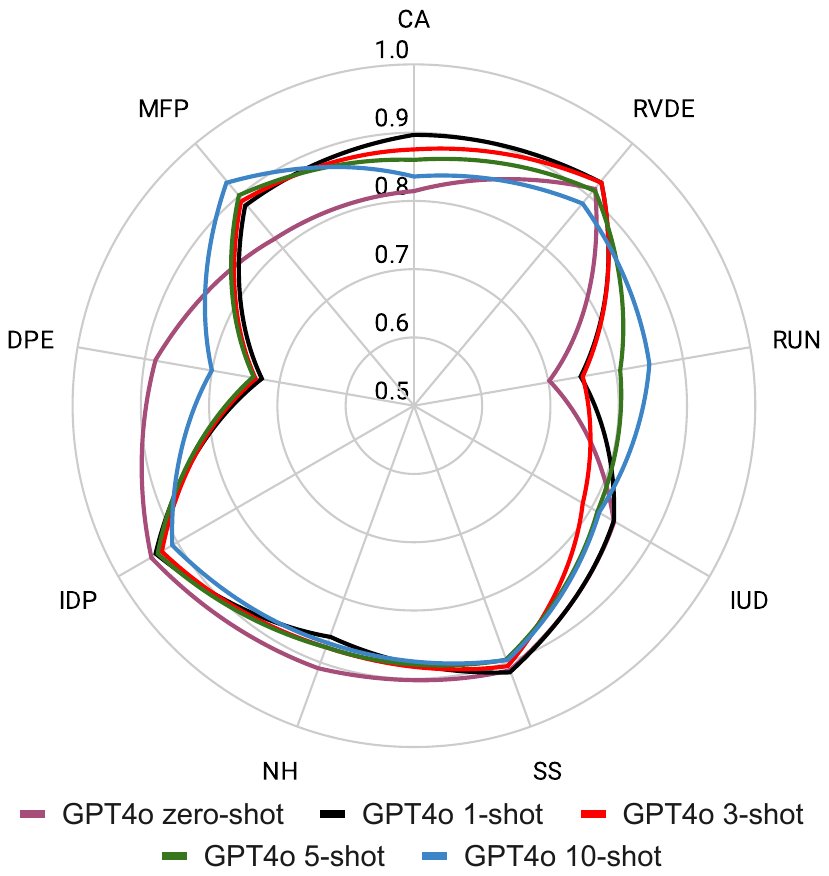}}\hspace{1cm}
    \subfloat[Mistral 7B in zero- and few-shot settings] 
    {\includegraphics[width=.40\linewidth]{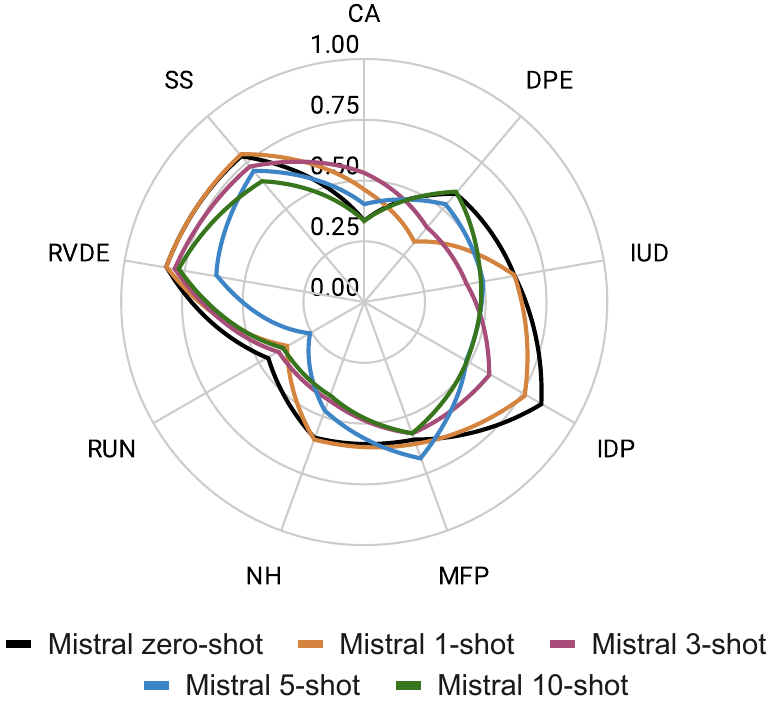}}
  \caption{Performance (F1 scores) of LLMs across various information types (i.e., classes)}
  \label{fig:information_type}
\end{figure*}

%\subsubsection{Proprietary models}
%\paragraph{Proprietary models}
Figure~\ref{fig:Disaster_wise}(a) shows the macro F1-scores for GPT-3.5, GPT-4, and GPT-4o across various few-shot settings. Notably, all models consistently show high performance for earthquakes, with GPT-4 achieving a maximum F1-score of 0.76 in the 10-shot setting and GPT-3.5 a minimum of 0.63 in the zero-shot setting. Conversely, model performances for floods consistently remain the lowest, with GPT-4o's 1-shot performance reaching the highest F1-score of 0.70, and GPT-3.5's zero-shot the lowest at 0.55. The results for wildfires and hurricanes are less consistent, though GPT-4o outperforms GPT-4 and GPT-3.5 in most cases. Surprisingly, increasing the number of shots does not show plausible performance improvements for all models. For GPT-3.5, there is a noticeable improvement from the zero-shot to other few-shot settings. However, for GPT-4, the performance from zero-shot to 3-shot remains nearly unchanged, and unexpectedly degrades in the 5-shot setting, and then recovers in 10-shot. Similarly, GPT-4o does not exhibit a consistent improvement as the number of shots increases.

%Among the proprietary models, earthquakes disaster type consistently outperformed in F1-score compared to the others. Our analysis reveals that in contrast to other disaster types, earthquakes contain \AW{proportionally} fewer classes that could potentially confuse the LLMs. For example earthquakes consists of 98 instances of {\it not humanitarian} and 45 instances of {\it requests or urgent needs}, while in contrast flood disaster type which has been consistently under-performed in F1-scores in all paid models contains 186 instances of {\it not-humanitarian} labels and 133 instances of {\it requests or urgent needs}. Upon doing class-wise grouping of the dataset in  for additional analysis in Figure \ref{fig:GPT4_information_type}, we note that {\it requests or urgent needs} and {\it not-humanitarian} are indeed most under-performing classes across all the proprietary models.

\begin{figure*}[t]
  \centering
  %\scalebox{0.90}{
   \subfloat[GPT-4o]
    {\includegraphics[width=0.44\linewidth]{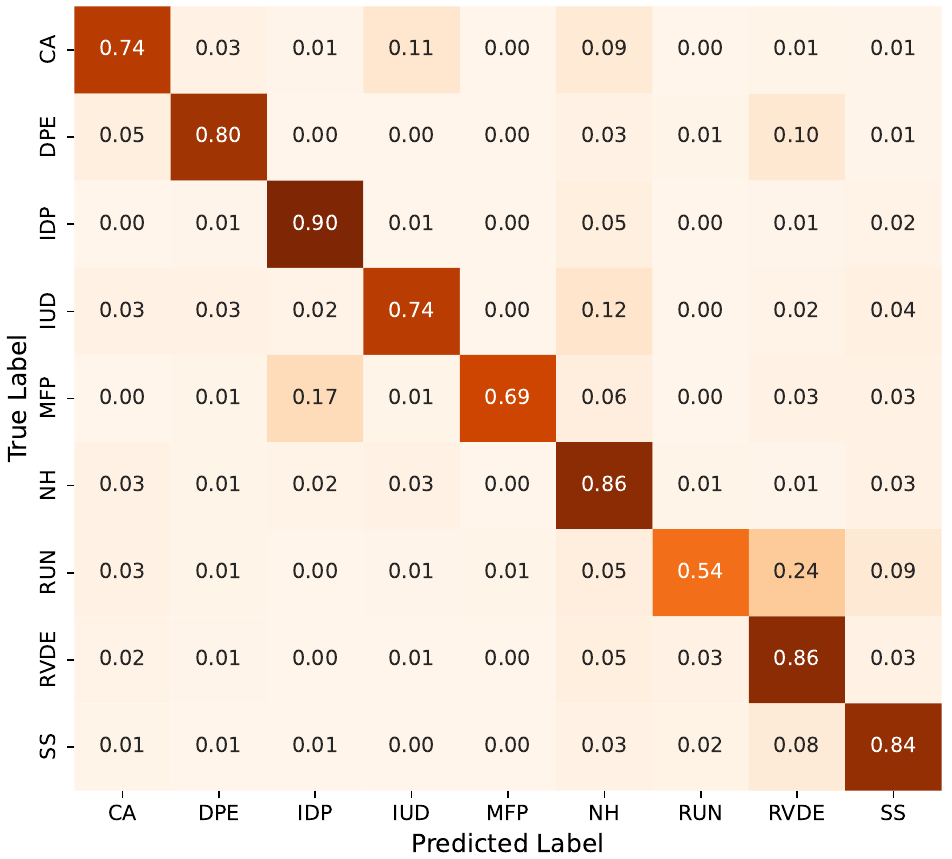} \hspace{1cm}}
    \subfloat[Mistral 7B] 
    {\includegraphics[width=.49\linewidth]{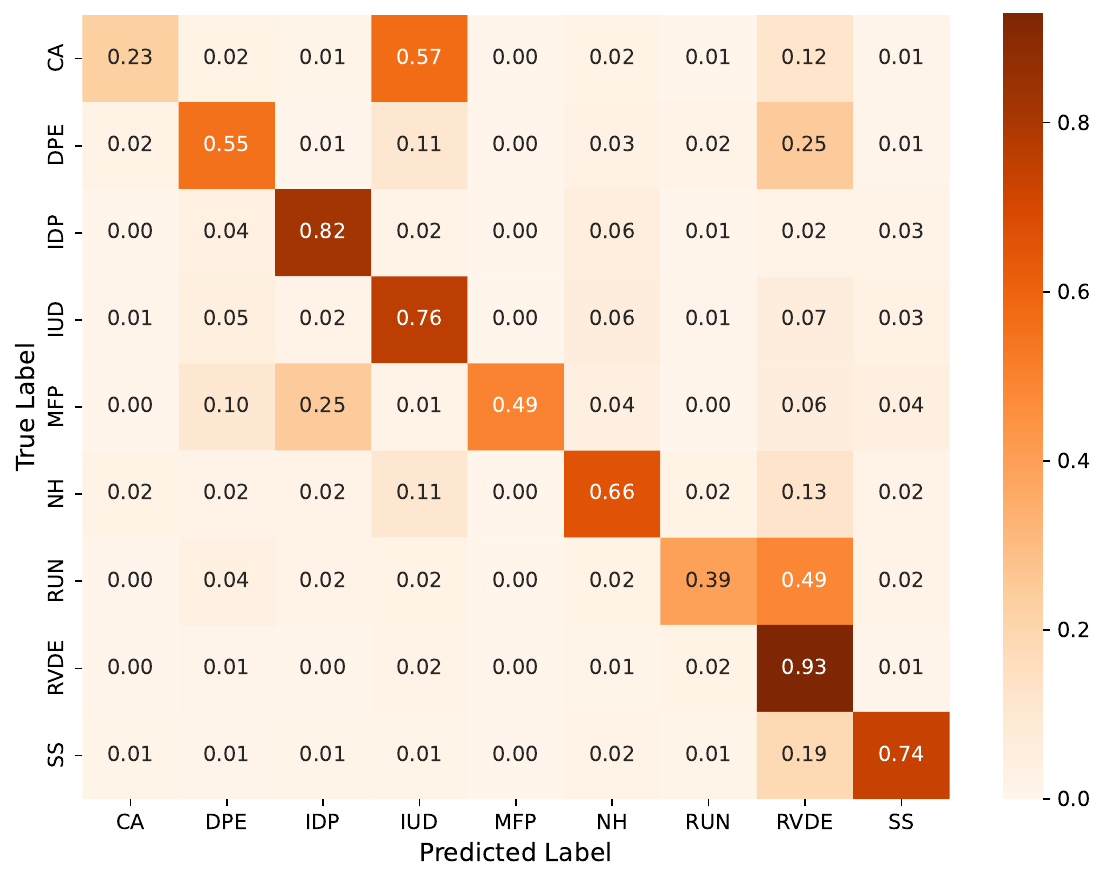}}
  \caption{Confusion matrices for GPT-4o (left) and Mistral 7B (right) models under the zero-shot setting}
  \label{fig:confusion_matrix}
\end{figure*}

% \subsubsection{Open-source models: Llama-2 \& Mistral}
% \begin{figure}[t]
%   \centering
%   \includegraphics[width=\linewidth]{figs/spider_open_Disaster.png} 
%   \caption{Open-Source Models Disaster Type Comparison}
%   \label{fig:open_disaster}
% \end{figure}

%\subsubsection{Open-source models}
%\paragraph{Open-source models}
Figure~\ref{fig:Disaster_wise}(b) presents the F1-scores for the Llama-2 13B, Llama-3 8B, and Mistral 7B models across various few-shot settings, excluding the Llama-2 10-shot due to token limit constraints. Overall, these open-source models perform less effectively than their proprietary counterparts. Specifically, Mistral's zero-shot achieves the highest F1-score of 0.62 for earthquakes and also shows similar results for hurricanes. Mistral consistently outperforms Llama-2 and Llama-3 across most cases. A notable observation is that the zero-shot setting generally yields the best results for both models, and adding more example shots does not significantly enhance performance. Overall, we observe that the open-source models tend to perform better for hurricanes as opposed to the proprietary models' superior performance for earthquakes.
%Unlike proprietary models, we do not observe any clear trend in open-source models across various disaster-types in Fig \ref{fig:Disaster_wise}(b). In general out of 5 Mistral variations, it performs well in 3 of its variations for  hurricanes disaster types, while 2 of its variations perform well in earthquakes. Mistral zero-shot is best performing model compared with its other variations as well as LLaMA, in overall disaster types. We therefore inspect its class-wise performance in Figure \ref{fig:Mistral_problematic_lables}, we observe that there is no distinct trend in identification of weak labels to attribute to performance variation of open-source models across different disaster types. Like properitery models, all variations of Mistral seem to be struggling in class labels: {\it requests or urgent needs}, {\it not-humanitarian} and {\it caution and advise} in Figure \ref{fig:Mistral_problematic_lables}. We shall further analyse class-wise performance of these models in Section \ref{sec:class-wise}.

\begin{figure*}[!htbp]
  \centering
  % Left figure
  \begin{minipage}{0.46\textwidth}
    \centering
    \includegraphics[width=\linewidth]{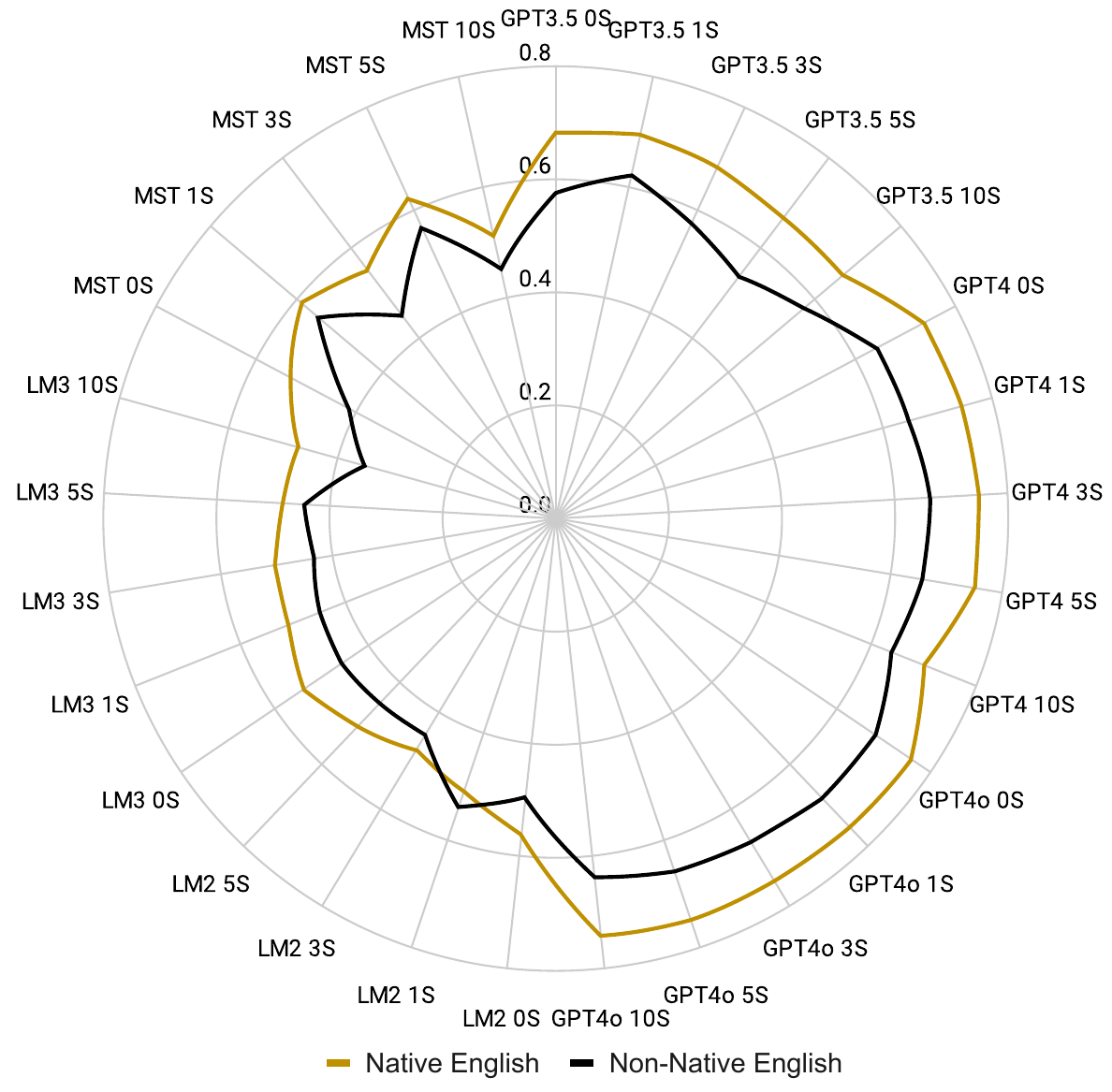}
    \caption{Performance (F1-scores) of LLMs on native-English-speaking vs.\ non-English-speaking countries. LM2=Llama-2 13B, LM3=Llama-3 8B, MST=Mistral 7B}
    \label{fig:native_non_native}
  \end{minipage}
  \hfill 
  % Right figure
  \begin{minipage}{0.53\textwidth}
    \centering
    \includegraphics[width=\linewidth]{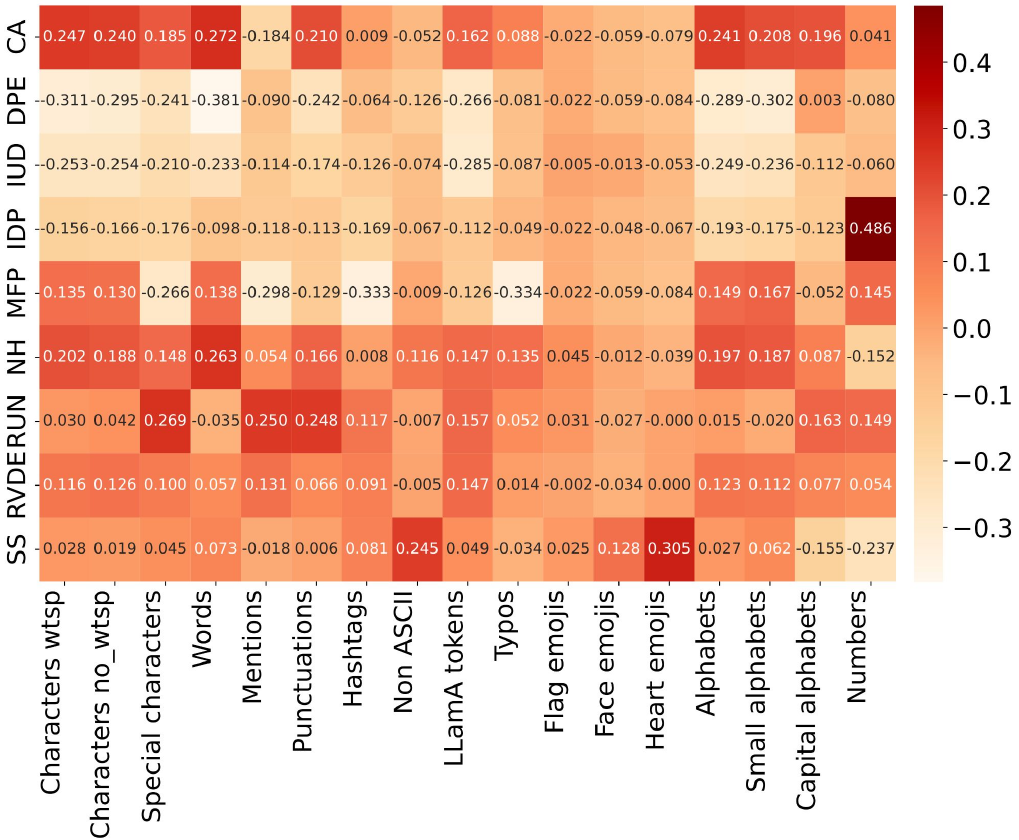} 
    \caption{Distribution of z-scores of language features (x-axis) across information classes (y-axis)}
    \label{fig:feature_heatmap}
  \end{minipage}
\end{figure*}

\subsection{Information Type Analysis}
\label{sec:class-wise}
Our second research question examines LLMs' capabilities in processing diverse types of information related to humanitarian response and situational awareness during disasters. Our analysis contains nine distinct information categories, detailed in Section~\ref{sec:Dataset}, with their cumulative distribution across all events depicted in Figure~\ref{fig:all_dist}(b). While we conducted experiments across all six models in all few-shot settings (except for Llama-2 10-shot), the following results focus solely on the two top-performing models, GPT-4o and Mistral, from the proprietary and open-source categories, respectively. The complete set of results, including all six models, are provided in Appendix~\ref{appendix:class-results}. 

%Our analysis in Sections \ref{sec:disaster-wise} \& \ref{sec:event-wise} revealed that the performance of LLMs to certain disaster and event types were linked to the \AW{quantity of problematic classes in that the disaster or event type, i.e if some disaster or an event had more of certain problematic class such as {\it requests and urgent needs} or {\it not humanitarian}, then it would under-perform in classification}. In this section therefore we analyse few case-by-case examples where these models failed {\AW{so that we can identify the underlying cause of either short-comings of the model to appropriately label these classes or short-comings of human-labellers to correctly attribute the tweets to designated classes}. 

% \begin{figure*}[t]
%   \centering
%     \subfloat[GPT-4 in zero and few-shot settings]
%     {\includegraphics[width=.30\linewidth]{new_figs/Spider_Class_GPTs.pdf}}\hspace{1cm}
%     \subfloat[Mistral in zero and few-shot settings] 
%     {\includegraphics[width=.30\linewidth]{new_figs/Spider_Class_OS.pdf}}
%   \caption{Performance (F1 scores) of LLMs across various information types (i.e., classes)}
%   \label{fig:information_type}
% \end{figure*}

%======

Figure~\ref{fig:information_type}(a) shows the class-wise macro F1-scores for GPT-4o models across all few-shot settings. These models consistently achieve F1-scores above 0.80 in all few-shot settings for the classes \textit{rescue volunteering or donation effort} (RVDE), \textit{sympathy and support} (SS) and \textit{injured or dead people} (IDP). In contrast, the \textit{requests or urgent needs} (RUN) and \textit{displaced people and evacuations} (DPE) classes consistently yield low performance, with F1-scores below 0.75, except for higher shots (i.e., 5 and 10). Notably, the \textit{requests or urgent needs} (RUN) class exhibits significant variability in performance across different shots. %, with the 0-shot being the lowest (F1=0.70) and the zero-shot the highest (F1=0.80). %To better understand the low performance in certain classes, we conducted an error analysis using the confusion matrix displayed in Figure~\ref{fig:confusion_matrix}, and selected random examples from the class \textit{``requests or urgent needs''} for further examination. We observed that general calls for  volunteering and donations in the tweets were confused by the model with the on-going effort for volunteering and donations which the class \textit{``rescue volunteering or donations effort''} represents. This explains why majority of tweets pertaining to \textit{``requests or urgent needs''} were misclassified as \textit{``rescue volunteering or donations effort''}. We believe that such subtle difference might even confuse some novice human annotators.
To understand why certain classes underperformed, we conducted an error analysis using the confusion matrix shown in Figure~\ref{fig:confusion_matrix}(a). We specifically examined random samples from the \textit{requests or urgent needs} (RUN)  class which are confused with the \textit{rescue volunteering or donations effort} (RVDE) class. Our analysis revealed that the model often confused general calls for volunteering and donations with ongoing volunteering efforts. This confusion led to a high rate of misclassification of tweets from \textit{requests or urgent needs} (RUN) as \textit{rescue volunteering or donation effort} (RVDE) (24\%), as shown in Figure~\ref{fig:confusion_matrix}(a).

Figure~\ref{fig:information_type}(b) presents the class-wise F1-scores of Mistral 7B across all few-shot settings. Mistral 7B notably under performs in the categories \textit{requests or urgent needs} (RUE) and \textit{caution and advice} (CA). Other instances of low performance include \textit{displaced people and evacuations} (DPE) in the 1-shot setting (F1=0.32), \textit{not humanitarian} (NH) in the 3-shot (F1=0.29), and most critically, \textit{requests or urgent needs} (RUE) in the 5-shot (F1=0.17). However, the model performs relatively well with \textit{rescue volunteering or donation effort} (RVDE) and \textit{injured or dead people} (IDP), especially in zero- and 1-shot scenarios. Overall, this open-source model lags behind its proprietary counterpart in information type classification performance. Figure~\ref{fig:confusion_matrix}(b) shows the confusion matrix of Mistral 7B 0-shot, which we used to perform an error analysis of mistakes made by the model. We observed that open-source models also confuse \textit{requests or urgent needs} (RUE) with  \textit{rescue volunteering or donations effort} (RVDE) due to the same reasoning where calls for volunteering or donations were mistaken with the efforts for volunteering or donations. %Moreover, we also reviewed several mistakes made by Mistral zero-shot on \textit{``caution and advice''} and noticed that whenever some intensity words such as \textit{``severe earthquakes''} or \textit{``category 5 hurricane''} appears in the tweets, the model potentially assumes some form of destruction and confuses with  \textit{``infrastructure or utility damage''}. This might be one of the reasons where significant portion of  \textit{``caution and advice''} tweets were misclassified as \textit{``infrastructure or utility damage''} as seen in Figure~\ref{fig:confusion_matrix}(b).
Additionally, we analyzed errors made by the Mistral zero-shot model in classifying \textit{caution and advice} (CA) tweets. We found that the presence of intensity descriptors such as ``severe earthquakes'' or ``category 5 hurricane'' led the model to mistakenly label tweets as \textit{infrastructure or utility damage} (IUD). The results from all the models are provided in Appendix~\ref{appendix:class-results}.

\begin{figure*}[t]
    \centering
    \begin{minipage}{0.60\textwidth}
        \centering
        \includegraphics[width=1\linewidth]{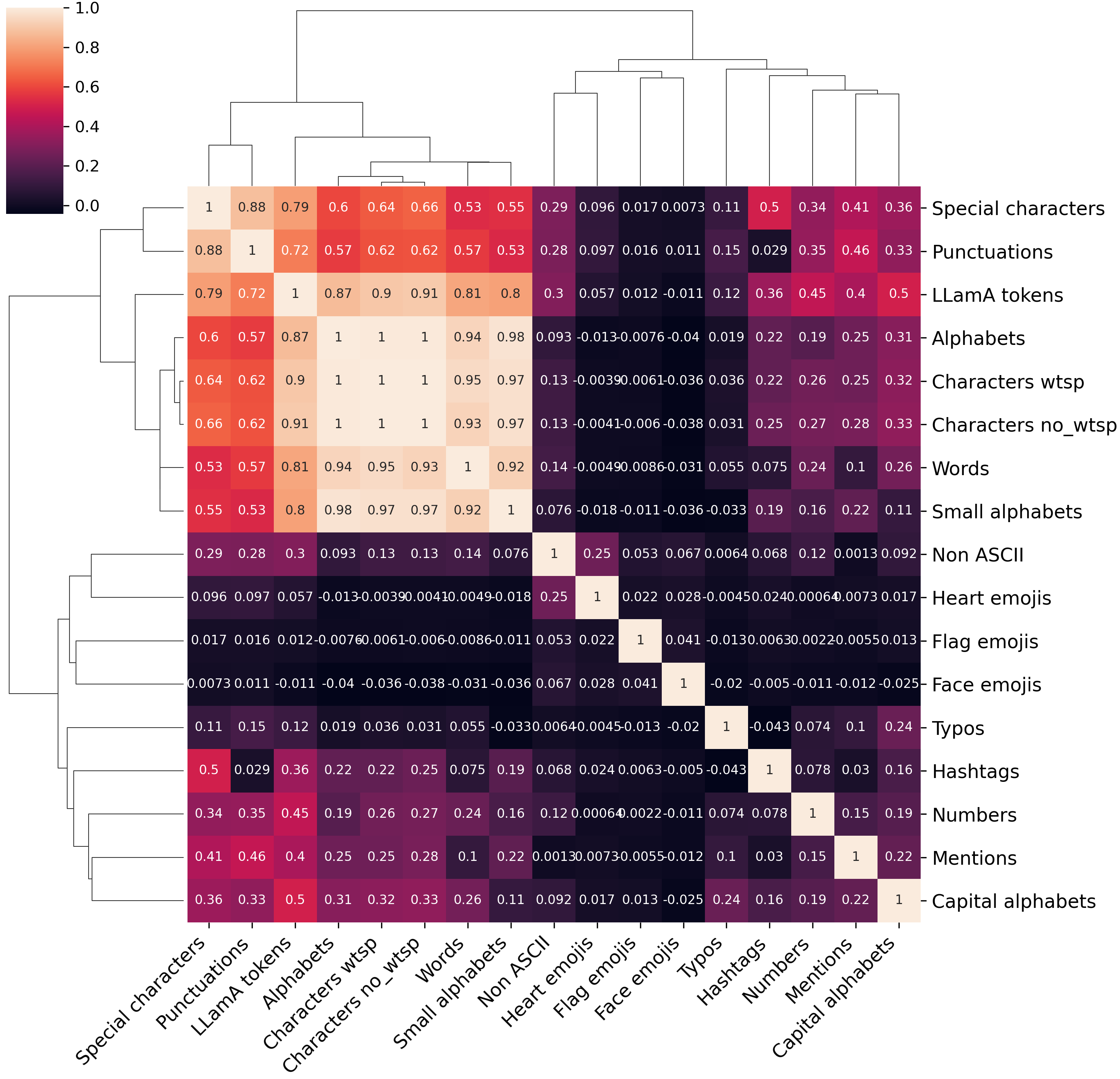} 
        \caption{Multicollinearity analysis of linguistic features}
        \label{fig:clustermap}
    \end{minipage}\hfill
    \begin{minipage}{0.40\textwidth}
        \centering
        \captionof{table}{Logistic regression analysis for Mistral-7B zero-shot}
  %\caption{Mistral zero-shot}
  \label{tab:regression_analysis_mistral}
  \scalebox{0.55}{
  \centering
  \begin{tabular}{@{}lrrrrrr@{}}
  \toprule
    & \multicolumn{1}{c}{coef} & \multicolumn{1}{c}{std err} & \multicolumn{1}{c}{z} & \multicolumn{1}{c}{P\textgreater{}|z|} & {[}0.025 & 0.975{]} \\ \midrule
  Intercept          &  1.3295 & 0.049 & 27.288 & 0.000 &  1.234 & 1.425 \\ 
  Typos              & -0.0457 & 0.029 & -1.586 & 0.113 & -0.102 & 0.011 \\ 
  Special characters & -0.0146 & 0.009 & -1.691 & 0.091 & -0.032 & 0.002 \\ 
  Characters         & -0.0037 & 0.000 & -9.562 & 0.000 & -0.004 & -0.003 \\ 
  Numbers       &  0.0239 & 0.006 &  3.737 & 0.000 &  0.011 & 0.036 \\
  Hashtags           &  0.0308 & 0.013 &  2.291 & 0.022 &  0.004 & 0.057 \\ 
  Mentions           &  0.0702 & 0.017 &  4.185 & 0.000 &  0.037 & 0.103 \\ 
  Face emojis        &  0.1462 & 0.224 &  0.651 & 0.515 & -0.294 & 0.586 \\ 
  Heart emojis       &  0.2719 & 0.143 &  1.905 & 0.057 & -0.008 & 0.552 \\ 
  \bottomrule
  \end{tabular}
  }

 \vspace{20pt} % Adjust the vertical space between the tables

\centering
\caption{Logistic regression analysis for GPT-4o zero-shot}
\label{tab:regression_analysis_gpt}
\scalebox{0.55}{
\centering
\begin{tabular}{@{}lrrrrrr@{}}
\toprule
    & \multicolumn{1}{c}{coef} & \multicolumn{1}{c}{std err} & \multicolumn{1}{c}{z} & \multicolumn{1}{c}{P\textgreater{}|z|} & {[}0.025 & 0.975{]} \\ \midrule
Intercept                &  1.6471 &  0.055 & 29.719 &  0.000 &   1.538 &   1.756 \\
Typos                    & -0.0575 &  0.033 & -1.768 &  0.077 &  -0.121 &   0.006 \\
Special characters       & -0.0115 &  0.010 & -1.171 &  0.241 &  -0.031 &   0.008 \\
Characters               & -0.0017 &  0.000 & -3.974 &  0.000 &  -0.003 &  -0.001 \\
Numbers                  &  0.0125 &  0.007 &  1.734 &  0.083 &  -0.002 &   0.027 \\
Hashtags                 &  0.0335 &  0.016 &  2.144 &  0.032 &   0.003 &   0.064 \\
Mentions                 &  0.0287 &  0.018 &  1.590 &  0.112 &  -0.007 &   0.064 \\
Face emojis              &  0.0996 &  0.254 &  0.391 &  0.695 &  -0.399 &   0.598 \\
Heart emojis             &  0.2166 &  0.164 &  1.324 &  0.185 &  -0.104 &   0.537 \\
\bottomrule
\end{tabular}
}
%\end{subtable}%
% \label{tab:regression_analysis}
%\end{table}
\end{minipage}
\end{figure*}
%=====

\begin{figure*}[!htbp]
  \centering
    \subfloat[Hashtags strictly at the beginning]
    {\includegraphics[width=.33\linewidth]{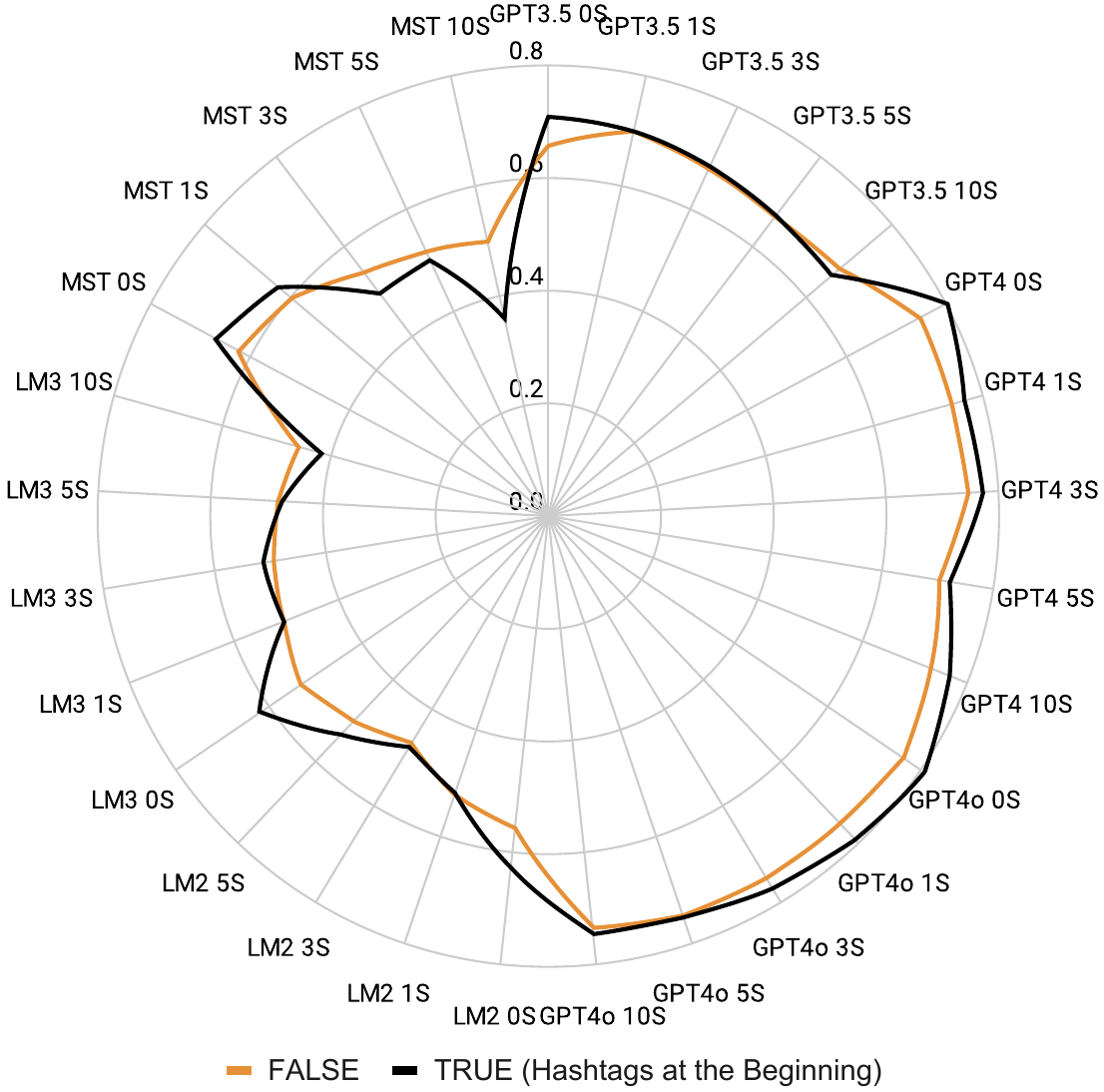}}\hfill
    \subfloat[Hashtags in the middle] 
    {\includegraphics[width=.33\linewidth]{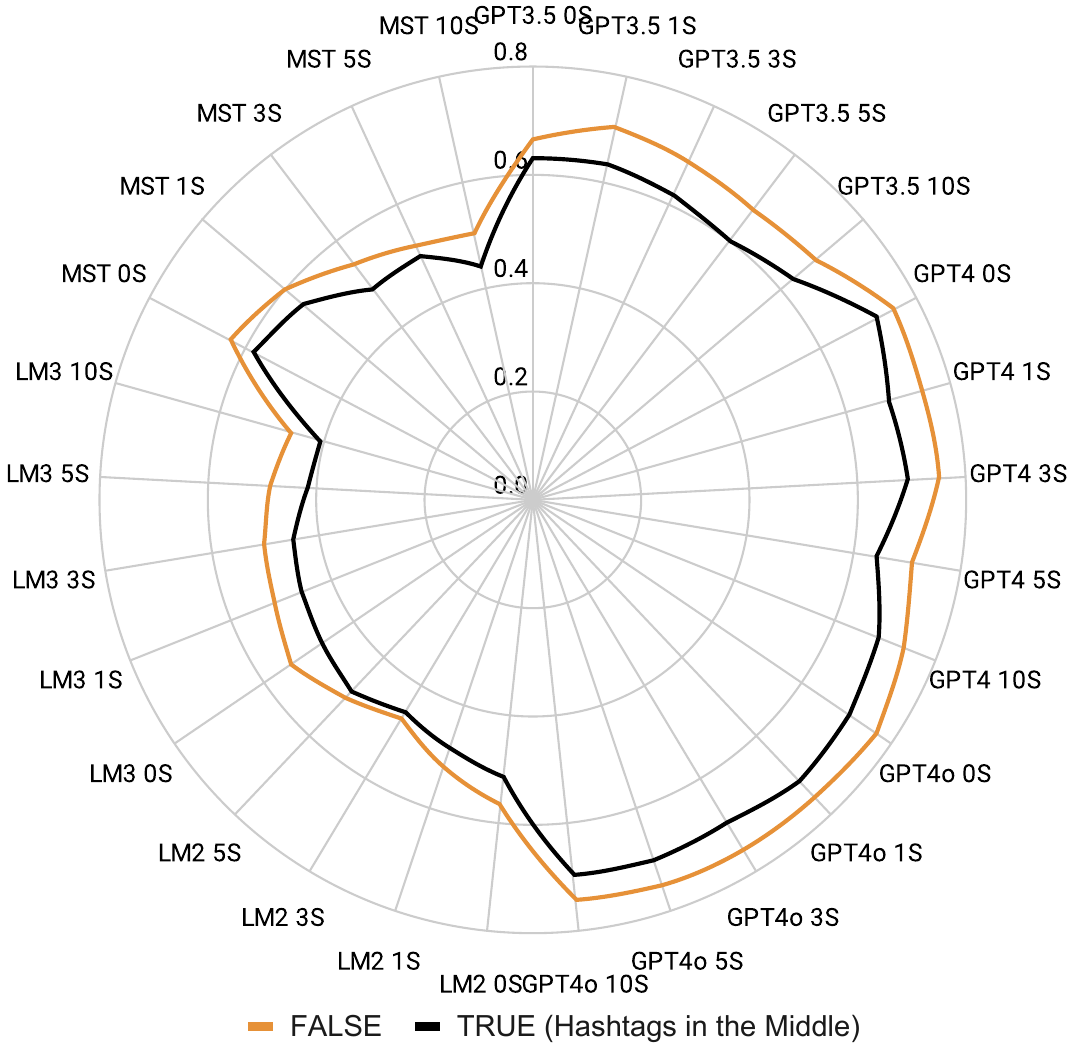}}\hfill
    \subfloat[Hashtags strictly at the end] {\includegraphics[width=.33\linewidth]{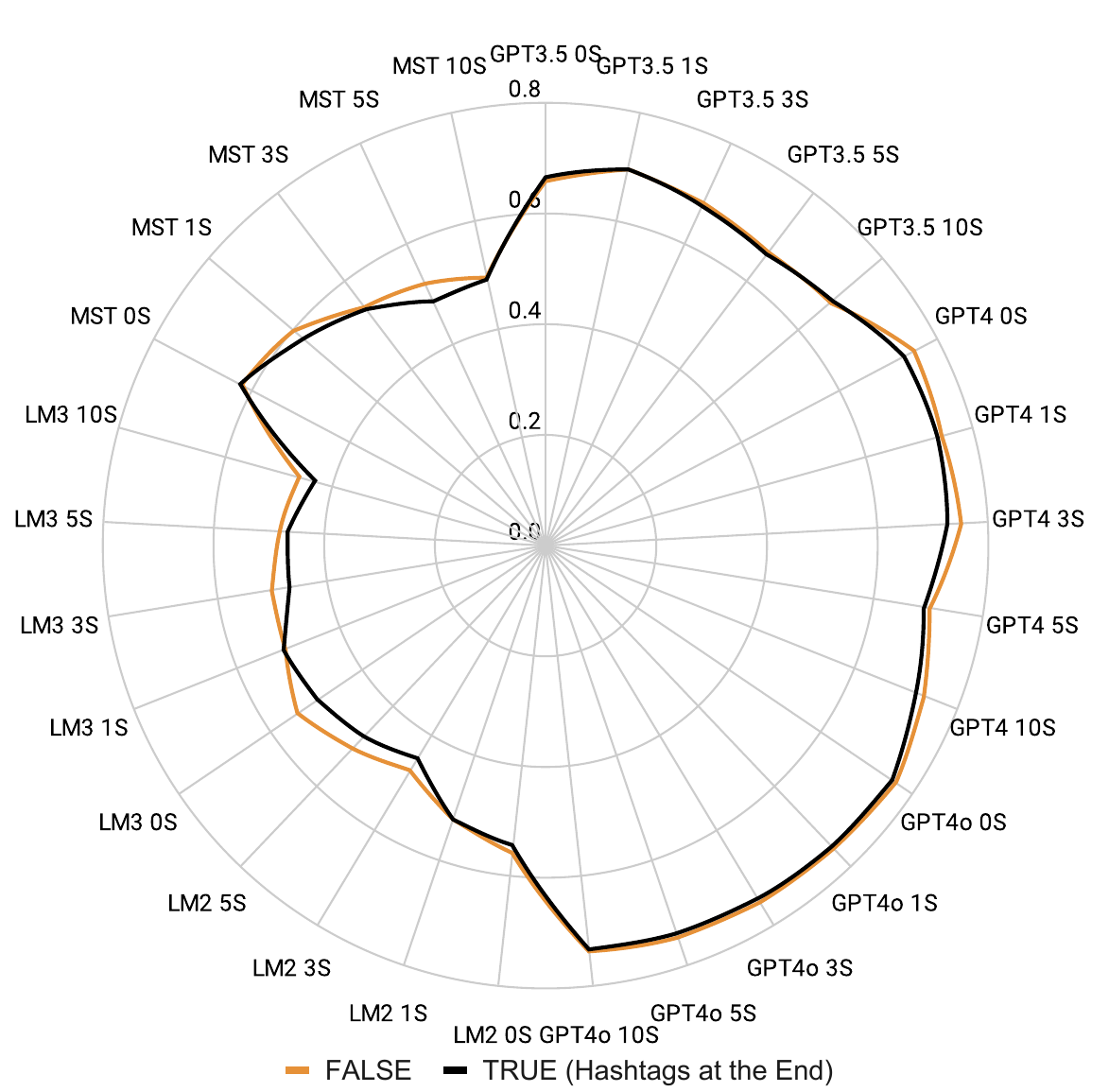}}
  \caption{Impacts of hashtag positioning on LLMs performance (macro F1-scores). LM2=Llama-2 13B, LM3=Llama-3 8B, MST=Mistral 7B}%True indicates the presence of hashtag condition within sentence and False indicates the opposite}
  \label{fig:hashtag_patterns}
\end{figure*}

%\AW{Additionally Mistral zero-shot is significantly miss-classifying labels belonging to  {\it caution and advise} with {\it infrastructure and utility damage}. This highlights significant misalignment as the representative examples of {\it caution and advise} are quite different then that of {\it infrastructure and utility damage} as seen in the their corresponding word cloud in Figure \ref{fig:word_cloud}. Our analysis reveal that any tweets pertaining to {\it caution and advise} often includes the reference of the disaster with the magnitude of the disaster, which Mistral potentially correlates to destruction and hence the label {\it infrastructure and utility damage}. }

\begin{figure}[!htbp]
  \centering
  \includegraphics[width=1\linewidth]{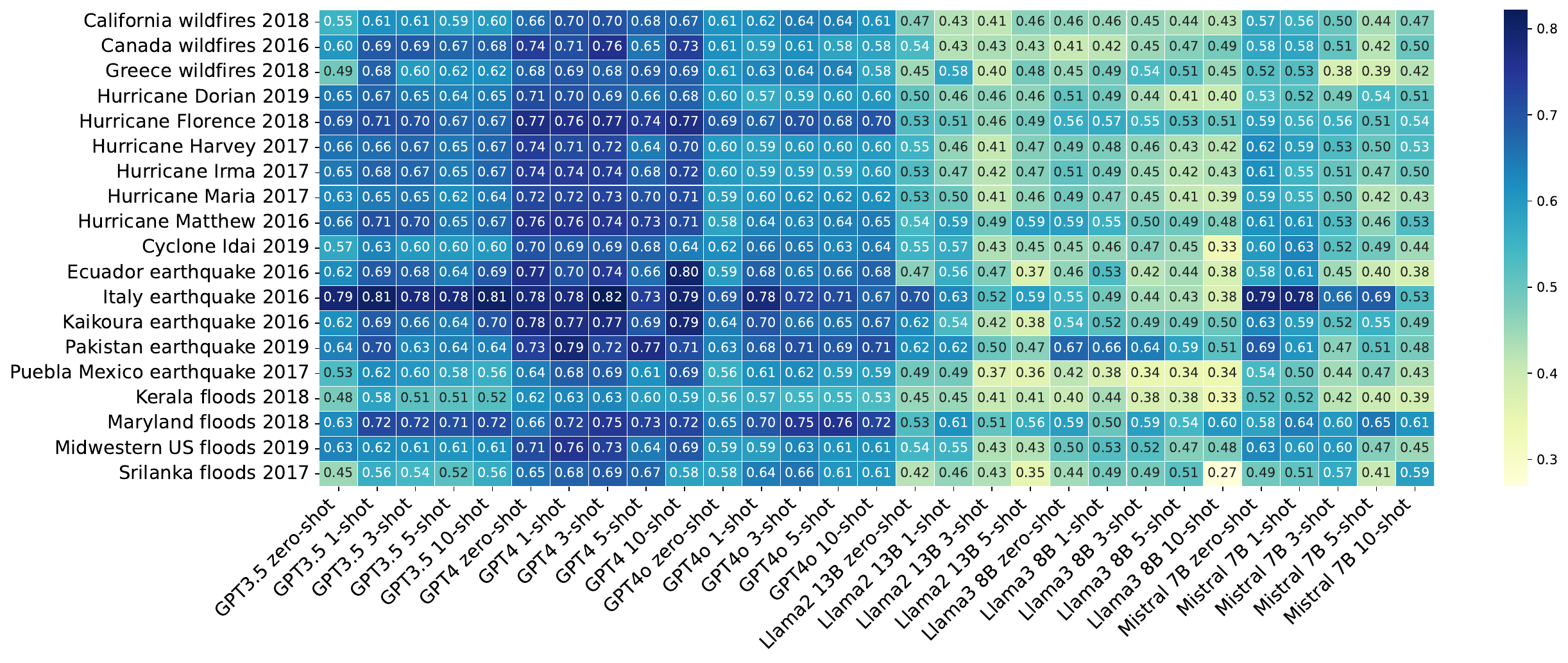} 
  \caption{F1-scores for all proprietary and open-source models for 19 events across all k-shot settings}
  \label{fig:event_wise}
\end{figure}

\subsection{Native vs.\ Non-Native English Analysis}
Our third research question examines the performance of LLMs in processing social media content from native-English-speaking versus non-English-speaking countries. Our dataset includes 11 events from native-English-speaking countries and 8 events from non-English-speaking countries. Figure~\ref{fig:all_dist}(c) illustrates the distribution of tweets across these two categories.

Figure~\ref{fig:native_non_native} displays the F1-scores for all models, including both proprietary and open-source. It is evident that all models achieve better performance in processing data from native-English-speaking countries. Proprietary models show a marked advantage in understanding data from these regions across all few-shot settings. However, their performance drops when processing data from non-English-speaking countries, although they still outperform open-source models for the same category. Furthermore, in the non-English-speaking category, GPT-4o zero-shot setting leads with an F1-score of 0.76, while Mistral in the 5-shot setting tops among open-source models with an F1-score of 0.62. The remaining open-source models generally score below 0.60, which is a surprising finding.% given that Mistral typically outperforms in most scenarios, as shown previously.

% \begin{figure}[t]
%   \centering
%   \includegraphics[width=0.50\linewidth]{new_figs/native.png} 
%   \caption{Performance (F1 scores) of LLMs on native vs.\ non-native English countries}
%   \label{fig:native_non_native}
% \end{figure}

\subsection{Linguistic Feature Analysis}
Our fourth research question explores whether various linguistic features, such as word count, hashtag count, and emoji usage in tweets, affect the performance of LLMs. Previous studies have shown that such features significantly influence the performance of traditional machine learning and deep learning models~\cite{castillo2011information}. We aim to determine if this holds true for LLMs, as well. We defined 17 linguistic features and analyzed their frequency distributions across all classes. Figure~\ref{fig:feature_heatmap} presents a heatmap of z-scores for these features' presence in each class, revealing notable patterns. For instance, the \textit{injured and dead people} (IDP) class has a high z-score for numbers, likely due to the prevalence of numerical data in such messages reporting casualties or injured people due to the disaster event. Similarly, the \textit{sympathy and support} (SS) class shows a high value for heart emojis, reflecting emotional expressions in such tweets. We observed that tweets discussing \textit{requests and urgent needs} often include more mentions of other user accounts, particularly NGOs and official accounts.

Next, we perform a logistic regression analysis to ascertain how different linguistic features affect model performance. To avoid the undesirable effects of multicollinearity, we exclude highly correlated linguistic features like character, word, and alphabet counts as illustrated in Figure~\ref{fig:clustermap} and work with a reduced set of features as our independent variables and consider the binary (correct/incorrect) validation of the predicted class labels with the ground truth as our dependent variable. Table~\ref{tab:regression_analysis_mistral} summarizes the analysis results for Mistral zero-shot. We see that numbers, hashtags, mentions, face, and heart emojis have positive correlations with model performance whereas character, special character and typo counts have the opposite effect. For example,---with a relatively small but statistically significant coefficient---, increasing character counts tends to negatively impact model performance. On the contrary,---again with relatively small but statistically significant coefficients---, number, hashtag, and mention counts play positive roles in improving predictive performance. According to Table~\ref{tab:regression_analysis_gpt}, these observations also hold for GPT-4o zero-shot model with the exception that coefficients for number and hashtag counts are not statistically significant as before. In both cases, it is notable that face and heart emojis have relatively larger coefficients and show more prominence in the regression analysis but lack statistical significance.

\paragraph{Hashtag Positioning Impacts}
Next, we investigate whether the placement of hashtags within messages affects LLM performance. We categorize messages into three groups: \textit{(i)} messages with hashtags only (strictly) at the beginning, \textit{(ii)} messages with hashtags in the middle, and \textit{(iii)} hashtags only (strictly) at the end. Figure~\ref{fig:hashtag_patterns} presents F1-scores for each scenario in separate radar charts. %Interestingly, we find that hashtags placed in the middle of messages often lead to more errors, as highlighted in Figure~\ref{fig:hashtag_patterns}(b) with the brown circle. This impact is more pronounced in proprietary models than in open-source models, as can be seen for GPT-3.5 all shots, except zero-shot, GPT-4, and GPT-4o. 
Interestingly, we observed that hashtags placed in the middle of messages frequently result in higher error rates, as illustrated in Figure~\ref{fig:hashtag_patterns}(b) with the brown circle. Most probably, these errors stem from the disruption in sentence structure caused by mid-sentence hashtags, which can confuse models by introducing unexpected breaks or context shifts. This phenomenon appears more pronounced in proprietary models compared to open-source models. Specifically, proprietary models such as GPT-3.5 (in all shot settings except zero-shot), GPT-4, and GPT-4o consistently exhibited difficulties in accurately interpreting messages with mid-sentence hashtags. The models often misclassify or overlook critical context surrounding the hashtag, leading to erroneous predictions.
Additionally, there is a notable difference in performance for GPT-4, GPT-4o, and Llama-3 zero-shot and Mistral 10-shot configurations when hashtags are exclusively at the beginning of messages. Conversely, the positioning of hashtags at the end of messages does not significantly affect LLMs' performance.

\subsection{Event-wise and Overall Performance}
In addition to our main analyses addressing the specified research questions, we conduct two additional experiments to assess LLMs' performance for individual events and their overall performance. These results also help benchmark the LLMs against this dataset.

Figure~\ref{fig:event_wise} shows the event-wise F1-scores for both proprietary and open-source models across various few-shot settings. Notably, the proprietary GPT models consistently outperform the open-source Llama and Mistral models. Among the proprietary models, all few-shot configurations of GPT-4 yield superior results compared to any few-shot setting of GPT-3.5 and GPT-4o. Specifically, GPT-4's zero-shot and 3-shot settings perform comparably and exceed the performance of its 5-shot and 10-shot settings. Interestingly, GPT-4o appears to face challenges in this experiment, particularly with hurricane and wildfire events. However, for earthquakes, GPT-4o's performance is comparable to or slightly below that of GPT-4. For GPT-3.5, the one-shot variant stands out as the most effective across the majority of events.

The event-wise results for open-source models (Figure~\ref{fig:event_wise}) highlight Mistral's zero- and one-shot settings as the most effective. In most cases, adding examples---increasing the number of shots---does not typically enhance the model's performance. Notably, larger number of shots, such as 5- or 10-shot, introduce additional tokens to the prompt, which may actually confuse the model rather than help it. However, both Llama 2 and 3 showed underwhelming performance across most events, with the exception of a few earthquake cases. In some instances, the Llama models scored as low as 0.27 (Sri Lanka floods) and 0.33 (Cyclone Idai).

Next, we evaluate the overall performance of LLMs on the entire data, including all events, information types, and language variations. Figure~\ref{fig:overall_f1} shows the F1-scores for both proprietary and open-source models across all shots and the SOTA supervised baseline (i.e., RoBERTa F1=0.78) as we report in \cite{alam2021humaid}. It is clear that GPT-4o consistently outperforms all other LLMs in all configurations, though it does not outperform the baseline. GPT-4 ranks as the second-best overall, while Llama-2 and Mistral generally underperform across all shots. Notably, there is no consistent trend in performance with the addition of more shots, with the exception of specific instances such as GPT-3.5's progression from zero to various few-shots, and Llama-2's improvement from 3- to 5-shot settings. We summarize the experimental results, including accuracy, precision, and recall of all the models across all shots in Table~\ref{tab:model_results}.

\begin{table}[t]
    \centering
    \begin{minipage}{0.55\textwidth}
        \centering
          % \includegraphics[width=\linewidth]{figs/event_wise_heatmap_all.pdf} 
          % \caption{F1-scores for all proprietary and open-source models across 19 events}
          % \label{fig:event_wise}
        % \includegraphics[width=\linewidth]{new_figs_gpt4o/Overall_F1_Scores.png} 
        \includegraphics[width=\linewidth]{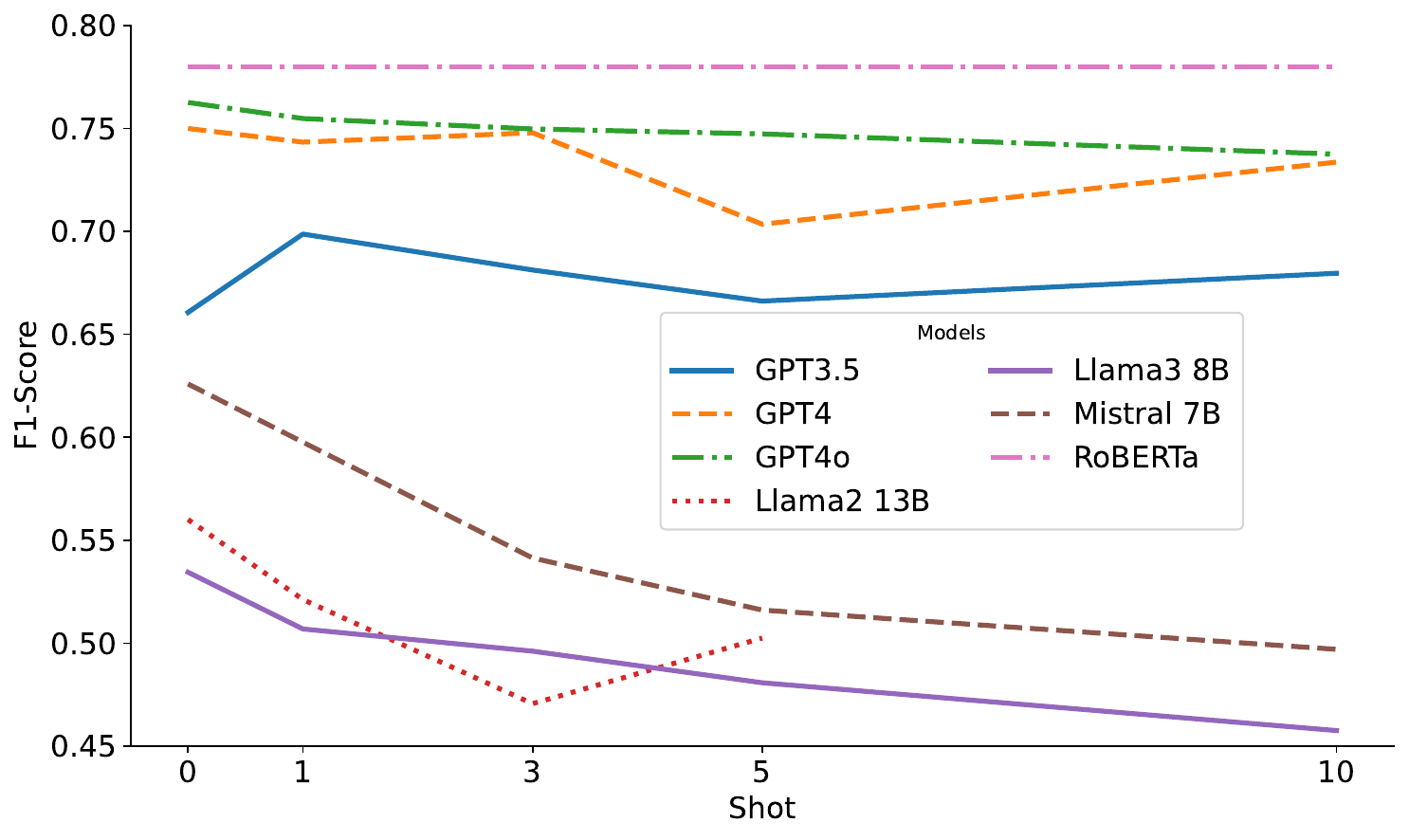} 
        \captionof{figure}{Overall performance of proprietary and open-source models across k-shot settings (k=\{0, 1, 3, 5, 10\}) and RoBERTa (F1=0.78) as a supervised baseline \cite{alam2021humaid}}
        \label{fig:overall_f1}
    \end{minipage}\hfill
    \vspace{5.5pt}
    \begin{minipage}{0.42\textwidth}
        \centering
        \caption{Comparison of LLMs' performance in terms of F1-score, Accuracy, Precision, and Recall}
        \label{tab:model_results}
        \resizebox{5cm}{!}{
        \begin{tabular}{llcccc}
        \toprule
        \# Shots & LLM Model                  & F1-score & Accuracy & Precision & Recall \\
        \midrule
        %\multicolumn{5}{c}{\textbf{Zero-shot}}\\
        %\midrule
        \multirow{6}{*}{0-shot} & GPT-4         & 0.750    & 0.785    & 0.764     & 0.747  \\
         & GPT-4o      & 0.762 & 0.801 & 0.771 & 0.760\\
         & GPT-3.5       & 0.661    & 0.686    & 0.729     & 0.644  \\
         & Llama-2 13B       & 0.562    & 0.554    & 0.694     & 0.522  \\
         & Llama-3 8B       & 0.534 & 0.540 & 0.621 & 0.547 \\
         & Mistral 7B      & 0.628    & 0.697    & 0.732     & 0.582  \\
        \midrule
         %\multicolumn{5}{c}{\textbf{1-shot}}\\
         %\midrule
        \multirow{6}{*}{1-shot} & GPT-4            & 0.743    & 0.769    & 0.777     & 0.728  \\
         & GPT-4o       & 0.755 & 0.80 & 0.763 & 0.760\\
         & GPT-3.5          & 0.699    & 0.748    & 0.733     & 0.675  \\
         & Llama-2 13B          & 0.522    & 0.522    & 0.655     & 0.559  \\
         & Llama-3 8B       & 0.507 & 0.520 & 0.603 & 0.532\\
         & Mistral 7B        & 0.598    & 0.682    & 0.702     & 0.563  \\
        \midrule
         %\multicolumn{5}{c}{\textbf{3-shot}}\\
         %\midrule
        \multirow{6}{*}{3-shot} & GPT-4            & 0.748    & 0.760    & 0.779     & 0.728  \\
        & GPT-4o       & 0.748 & 0.787 & 0.766 & 0.748\\
         & GPT-3.5          & 0.681    & 0.729    & 0.718     & 0.666  \\
         & Llama-2 13B          & 0.471    & 0.430    & 0.660     & 0.508  \\
         & Llama-3 8B       & 0.496 & 0.518 & 0.620 & 0.551 \\
         & Mistral 7B        & 0.543    & 0.592    & 0.652     & 0.526  \\
        \midrule
         %\multicolumn{5}{c}{\textbf{5-shot}}\\
         %\midrule
        \multirow{6}{*}{5-shot} & GPT-4            & 0.703    & 0.726    & 0.756     & 0.685  \\
        & GPT-4o       & 0.747 & 0.784 & 0.759 & 0.758\\
         & GPT-3.5          & 0.666    & 0.715    & 0.719     & 0.638  \\
         & Llama-2 13B          & 0.504    & 0.457    & 0.644     & 0.513  \\
         & Llama-3 8B       & 0.481 & 0.498 & 0.623 & 0.545 \\
         & Mistral 7B        & 0.516    & 0.513    & 0.614     & 0.531  \\
        \midrule
         %\multicolumn{5}{c}{\textbf{10-shot}}\\
         %\midrule
        \multirow{5}{*}{10-shot}  & GPT-4           & 0.734    & 0.730    & 0.779     & 0.702  \\
        & GPT-4o       & 0.737 & 0.769 & 0.744 & 0.764 \\
         & GPT-3.5         & 0.680    & 0.729    & 0.721     & 0.660  \\
         & Llama-3 8B       & 0.457 & 0.463 & 0.580 & 0.512\\
         & Mistral 7B       & 0.521    & 0.556    & 0.599     & 0.523  \\
        \bottomrule
        \end{tabular}
        }
    \end{minipage}
\end{table}

%% file: conclusion.tex
\section{Ethical Considerations}
The datasets used in this study consist of publicly available tweets posted by individuals or organizations during various natural disasters. The data was collected in strict adherence to the terms and conditions set forth by the Twitter (now X) API to ensure ethical compliance. To safeguard individuals' privacy, any personally identifiable information, including names, addresses, phone numbers, or other sensitive details, was systematically anonymized before data processing. Moreover, no attempts were made to infer or store additional demographic or personal information about the users. 

\section{Conclusion and Future Work}
\label{sec:conclusion}
We presented a comprehensive evaluation of prominent large language models in processing social media data from 19 major natural disasters across 11 countries, including 8 non-native and 3 native English-speaking regions. Our findings highlight varying strengths and limitations of LLMs in managing diverse disaster types, information categories, and linguistic complexities. Specifically, the models demonstrated notable difficulties with flood-related data and frequently misclassified critical information categories such as \textit{requests and urgent needs} and \textit{caution and advice}. Furthermore, our analysis identified key factors such as message length, typographical errors, and the presence of special characters as significant challenges that impair model performance. Importantly, we observed that providing few-shot examples yielded limited performance gains for most models. This could be due to the high variability in social media content, even from the same class. Finally, we provided benchmarking results, aiming to inform further research into LLMs' vulnerabilities and assist in developing more robust models for disaster information processing. \\

%By offering benchmarking results, this study contributes to the growing body of research aimed at improving LLM performance in disaster information processing. The insights provided here can serve as a foundation for designing more robust models that better support crisis response and management.

%We presented a detailed analysis of well-known LLMs to evaluate their performance on social media data from 19 major natural disasters across 15 countries, including 11 non-native and 4 native English-speaking. Our findings highlight both strengths and weaknesses of LLMs in handling different disaster types, information categories, and linguistic features. Notably, both proprietary and open-source models struggled with flood-related data and frequently misinterpreted critical information categories such as \textit{requests and urgent needs} and \textit{caution and advice}. Our analysis also revealed that message length, typos, and special characters negatively affect model performance. Additionally, providing example shots to models showed minimal improvement. Finally, we provided benchmarking results, aiming to inform further research into LLMs' vulnerabilities and assist in developing more robust models for disaster information processing. \\
%{\bf Limitations:} This study \textit{(i)} lacks data from human-induced events such as wars and conflicts; \textit{(ii)} not covering all natural disasters, missing types like droughts, and blizzards; \textit{(iii)} not including the most recent LLMs, such as GPT-4o and Llama-3; and \textit{(iv)} is limited to classification tasks only.\\

\noindent {\bf Future work:} 
We aim to extend our qualitative analyses to understand the reasons behind LLMs' underperformance for specific disaster types and information categories, with a focus on identifying actionable solutions to address these issues. %Additionally, we plan to evaluate the performance of emerging LLMs, such as GPT-4o and LLaMA-3, on similar datasets to benchmark advancements in the field. 
Beyond text-based models, our future research will explore the potential of large vision-language models in processing multimodal social media data, such as combining textual and visual content, to provide a more holistic understanding of disaster events. This exploration is particularly relevant for enhancing emergency management systems in complex real-world scenarios.